%% file: main.tex
\documentclass[runningheads]{llncs}

\usepackage{eccv}

\usepackage{eccvabbrv}

\usepackage{graphicx}
\usepackage{booktabs}

\usepackage{bm}
\usepackage{amsmath}
\usepackage{booktabs}   % for \toprule, \midrule, \bottomrule, \cmidrule
\usepackage{multirow}   % for \multirow
\usepackage{siunitx}
\usepackage{makecell}

\usepackage[accsupp]{axessibility}  % Improves PDF readability for those with disabilities.

\usepackage{hyperref}

\usepackage{orcidlink}
\usepackage{amsmath}

\usepackage{algorithm}
\usepackage{algpseudocode}

\begin{document}

% \title{Author Guidelines for ECCV Submission} 
\title{SAFE-Pruner: Semantic Attention–Guided Future-Aware Token Pruning for Efficient Vision-Language-Action Manipulation}

\titlerunning{SAFE-Pruner}

\author{Shilin Ma\inst{1} \and
Chubin Zhang\inst{1} \and
Changyuan Wang\inst{1} \and
Yuji Wang\inst{1} \and
Yue Wu\inst{1} \and \\
Zixuan Wang\inst{1} \and
Jingqi Tian\inst{1} \and
Zheng Zhu\inst{2} \and
Yansong Tang\inst{1}$^\dagger$}

\authorrunning{S. Ma et al.}

% \institute{
% Tsinghua Shenzhen International Graduate School, Tsinghua University \and
% GigaAI           
% }
% \institute{%
% $^{1}$ Tsinghua Shenzhen International Graduate School, Tsinghua University
% \quad
% $^{2}$ GigaAI
% }

\institute{
Tsinghua Shenzhen International Graduate School, Tsinghua University, China \and
GigaAI, China \\
 \email{\{msl21@mails, tang.yansong@sz\}.tsinghua.edu.cn}
}

\maketitle

\begingroup
\renewcommand{\thefootnote}{\ensuremath{\dagger}}
\footnotetext{Corresponding author}
\endgroup
\setcounter{footnote}{0}

\input{sections/0_abstract}
\input{sections/1_introduction}
\input{sections/2_related_work}
\input{sections/3_method}
\input{sections/4_experiment}
\input{sections/5_conclusion}

\section*{Acknowledgements}
% This work was supported by Shenzhen Science and Technology Program (JCYJ20240813111903006).

This work was supported by Shenzhen Science and Technology Program under Grant JCYJ20240813111903006.

\bibliographystyle{splncs04}
\bibliography{main}

% \appendix
% \input{sections/X_supp}

\end{document}

%% file: sections/0_abstract.tex
\begin{abstract}
Real-time inference of vision-language-action (VLA) models is essential for robotic control.  While visual token pruning has shown strong potential for accelerating inference, most existing methods mainly base pruning decisions on shallow-layer cues and risk discarding visual information required by deep layers. To address this issue, we propose \textbf{SAFE-Pruner}, a plug-and-play pruning framework that incorporates attention cues of future layers into pruning decisions. Specifically, we identify semantic attention consistency, the tendency that VLA models concentrate their attention probability mass on the same semantic entity across control timesteps. Based on this observation, we design a forward-looking strategy to forecast the token saliency in deep layers, which prevents the premature removal of critical tokens and leads to more stable acceleration. We further introduce a reference timestep refresh strategy that triggers updates upon attention shifts, thereby improving forecasting accuracy and pruning reliability. Extensive experiments across diverse evaluation settings demonstrate that our method achieves up to $1.89\times$ speedup with a minimal degradation in success rate of less than 1.5\%, while outperforming state-of-the-art methods by up to 1.9\%.
\href{https://msssl.github.io/SAFE-Pruner/}{\textcolor{blue}{Project page.}}
  \keywords{Vision-Language-Action Model \and Token Pruning \and Inference Acceleration}
\end{abstract}

%% file: sections/1_introduction.tex
\section{Introduction}
\label{sec:introduction}

Vision-Language-Action (VLA) models~\cite{kim2024openvla,kim2025openvlaoft,li2024cogact,black2024pi_0,intelligence2025pi05visionlanguageactionmodelopenworld,bjorck2025gr00t} have become a promising route toward general and flexible robotic manipulation, building on the advances in pretrained vision-language backbones~\cite{karamcheti2024prismatic,beyer2024paligemmaversatile3bvlm,bai2025qwen25vltechnicalreport} and large-scale robotic datasets~\cite{dasari2020robonetlargescalemultirobotlearning,kalashnikov2021mt,ebert2021bridgedataboostinggeneralization,mees2022calvin,choi2024lotabenchbenchmarkinglanguageorientedtask}. By unifying visual perception~\cite{zhu2026segment,tian2025ddavsdisentangledaudiosemantics,wang2025dynamic}, language understanding~\cite{wang-etal-2025-ponder,yang2022lavt} and downstream generation~\cite{zhu2025kv,zhu2025fade,zhu2025memorize} in a single framework, these models enable end-to-end embodied control across diverse tasks and environments. However, the heavy computational overhead and high inference latency severely hinder their application for real-time robotic control, especially in dynamic scenarios where timely responsiveness is critical. Thus, developing lightweight acceleration methods for VLA models is an urgent and critical research challenge for enabling real-world robotic perception and control.

While traditional acceleration techniques such as distillation~\cite{dong2025vitavlaefficientlyteachingvisionlanguage,zhuvarestorer}, quantization~\cite{xu2026qvlachannelsequalvisionlanguageaction,lv2026enhancingposttrainingquantizationfuture, wang2024q} and layer pruning~\cite{zhang2025molevladynamiclayerskippingvision} provide efficiency gains for VLA models, their requirement for retraining or architectural modifications severely limits their generalization capability. Recently, visual token pruning~\cite{chen2024fastv,zhang2024sparsevlm,alvar2025divprune,xu2025vlacache,liu2026vlaprunertemporalawareduallevelvisual} has become a promising direction, which reduces computation by discarding visually redundant tokens based on attention probabilities~\cite{chen2024fastv,zhang2024sparsevlm} or similarity scores~\cite{alvar2025divprune,zhang2025attentionsimilaritymaximizingconditional}. Some methods further integrate VLA-specific properties into pruning decisions to enhance stability in long-horizon control~\cite{xu2025vlacache,pei2025actionawaredynamicpruningefficient}. 

Despite these advances, current token pruning methods for VLA models face a short-sighted selection issue. Most existing approaches~\cite{chen2024fastv,zhang2024sparsevlm,alvar2025divprune,liu2026vlaprunertemporalawareduallevelvisual} make pruning decisions using signals extracted from early intra-timestep stages like shallow-layer attention, and then permanently discard the remaining tokens, which fails to fully account for their downstream utility within the same control timestep. In VLA models, producing a single action typically involves multi-stage computation like layer-wise feature refinement from shallow to deep Transformer layers; as a result, tokens that appear less salient early on may later become informative for deep-layer reasoning and prematurely pruning such tokens can therefore cause significant performance degradation. While recent works~\cite{liu2026vlaprunertemporalawareduallevelvisual,wang2025specprunevla} attempt to address this issue, they mainly estimate late-stage token saliency from visual-token subsets already pruned in early layers, where some critical tokens may have been irreversibly removed.
\input{figures/comparison}

To address the above limitations, we propose SAFE-Pruner, a plug-and-play framework that incorporates token saliency of different inference stages into pruning decisions to avoid premature discarding of critical tokens. Our goal is to predict intra-timestep future reasoning cues during early pruning stages based on historical frames. To achieve this goal, we first analyze how the attention scores of VLA models evolve across different frames within the same rollout.  Specifically, we identify a phenomenon of semantic attention consistency: when performing the same task across different timesteps, the model predominantly attends to patches with consistent semantic information, such as target objects to be grasped, even when their spatial locations change. Based on this phenomenon, we propose a future attention forecast strategy that uses late-stage inference information from historical frames to predict future reasoning cues of the current frame, thereby avoiding the premature pruning of tokens that will be attended to in late stages and leading to consistent performance gains across diverse benchmarks and tasks. While semantic attention consistency is well-maintained most of the time, we still observe distinct changes in attention patterns at the boundaries between two significantly different subtasks, which degrade the prediction accuracy. In addition, long-horizon forecasting may accumulate drift over time. To address these issues, we propose an adaptive key timestep selection strategy based on early-stage attention changes, which refreshes the reference when necessary to maintain consistency between the current and reference timesteps and avoid interference from outdated information. Fig.~\ref{fig:comparison} provides a visual comparison of our method against conventional shallow-only approaches.

We conduct extensive and well-designed experiments to validate the effectiveness and generalization capability of our method. Specifically, we implement our approach alongside state-of-the-art open-source methods~\cite{chen2024fastv,zhang2024sparsevlm,alvar2025divprune,xu2025vlacache,liu2026vlaprunertemporalawareduallevelvisual} across four distinct VLA architectures~\cite{kim2024openvla,kim2025openvlaoft,intelligence2025pi05visionlanguageactionmodelopenworld,li2024cogact} and benchmarks~\cite{liu2023libero,li24simpler}. The experimental results demonstrate that our method outperforms existing approaches in both inference acceleration and task performance.

Our main contributions can be summarized as follows: 
\begin{itemize}
    \item We identify and validate the semantic attention consistency phenomenon of VLA models in continuous manipulation, revealing a critical attention regularity for pruning decision-making; 
    \item We propose a future-aware token pruning framework with an attention prediction mechanism based on semantic consistency and adaptive key timestep selection, mitigating the short-sighted pruning problem of existing methods;
    \item We conduct comprehensive evaluations in diverse settings, demonstrating that our method achieves up to $1.89\times$ speedup with a negligible drop in success rate below 1.5\%, while outperforming state-of-the-art methods by up to 1.9\%.
\end{itemize} 

%% file: figures/comparison.tex
\begin{figure}[tb]
  \centering
  \includegraphics[width=0.99\linewidth]{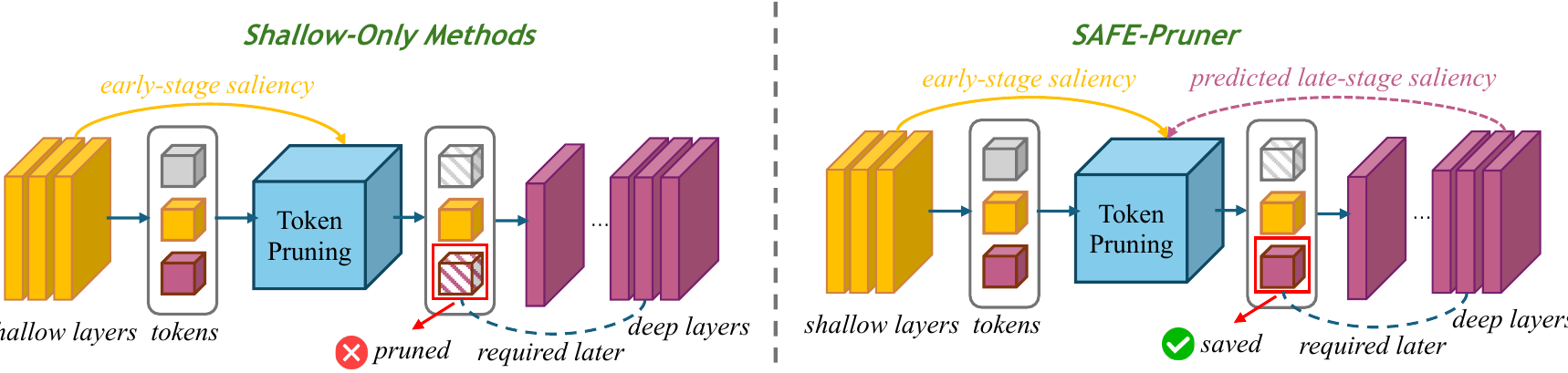}
  \caption{\small \textbf{Comparison of SAFE-Pruner and shallow-only token pruning methods.} Shallow-only methods (left) rely on early-stage saliency, risking premature pruning of tokens critical for deep layers. SAFE-Pruner (right) comprehensively considers saliency across inference stages by predicting late-stage importance, enabling correct retention of important tokens and achieving a better accuracy-efficiency balance.}
  \label{fig:comparison}
\end{figure}

%% file: sections/2_related_work.tex
\section{Related Work}
\label{related_work}

\noindent\textbf{Vision-Language-Action Models (VLA).}
\quad Vision-language-action (VLA) models aim to unify visual perception, language understanding, and action generation within a single policy for embodied manipulation~\cite{driess2023palmeembodiedmultimodallanguage,brohan2023rt1roboticstransformerrealworld,brohan2023rt2visionlanguageactionmodelstransfer,reed2022generalistagent}. A common recipe is to leverage a pre-trained vision-language model (VLM)~\cite{pmlr-v139-radford21a,chen2023palixscalingmultilingualvision} and scale learning with large collections of robot demonstrations~\cite{brohan2023rt1roboticstransformerrealworld,o2024openx-embodiment,dasari2020robonetlargescalemultirobotlearning}, which enables open-vocabulary instruction following and improved generalization across tasks, objects, and environments. Recent work further explores different action decoding formulations, which include autoregressive policies~\cite{kim2024openvla,zhang2026clap}, action-chunk prediction~\cite{zhao2023learningfinegrainedbimanualmanipulation,kim2025openvlaoft}, and diffusion-style policy heads~\cite{chi2025diffusionpolicy,li2024cogact}, as well as data-centric strategies such as multi-task training~\cite{bjorck2025gr00t,intelligence2025pi05visionlanguageactionmodelopenworld}. Together, these advances establish VLA as a promising paradigm for general-purpose manipulation, while also bringing increased computational demands due to dense visual tokens and long-horizon, closed-loop inference~\cite{yang2025efficientvla}.

\noindent\textbf{Acceleration for VLA Models.}
\quad Improving the inference efficiency of VLA models is important for real-time robotic control, where latency can directly affect closed-loop stability and task success. Acceleration efforts span both system- and model-level directions, comprising both general-purpose techniques and VLA-specialized optimizations. On the system side, prior work improves throughput using mixed-precision inference~\cite{micikevicius2018mixed}, operator/kernel optimizations~\cite{dao2022flashattention}, and better KV-cache management~\cite{kwon2023efficientmemorymanagementlarge}. On the model side, researchers study lightweight architectures~\cite{dong2025vitavlaefficientlyteachingvisionlanguage,zhang2025molevladynamiclayerskippingvision}, as well as training-free methods~\cite{yue2024deer} that adaptively reduce computation. In particular, visual token pruning~\cite{chen2024fastv,liu2026vlaprunertemporalawareduallevelvisual} is attractive in VLA settings because visual tokens typically dominate sequence length and attention cost. By selecting a compact subset of informative patches guided by attention features~\cite{chen2024fastv,pei2025actionawaredynamicpruningefficient,wang2025specprunevla} or similarity signals~\cite{alvar2025divprune}, these methods can yield substantial speedups with limited performance degradation. Nevertheless, pruning at high ratios remains challenging, as token importance may shift across layers and different stages, motivating approaches that better anticipate downstream cues and preserve task-critical visual evidence to achieve better performance.

%% file: sections/3_method.tex
\section{Method}
\subsection{Preliminary: Attention-based Token Saliency}

Current works~\cite{chen2024fastv,zhang2024sparsevlm,xu2025vlacache,liu2026vlaprunertemporalawareduallevelvisual} typically utilize the cross-attention weights to identify the important tokens and guide the pruning.  
Let $\ell_s$ denote the layer at which token pruning is performed, $\mathbf{H}^{\ell_s}_{\mathrm{vis}} \in \mathbb{R}^{L_{\mathrm{vis}} \times D}$ and $\mathbf{H}^{\ell_s}_{\mathrm{query}} \in \mathbb{R}^{L_{\mathrm{query}} \times D}$ represent the hidden state matrices of visual tokens and query tokens respectively, where $L_{\mathrm{vis}}$ and $L_{\mathrm{query}}$ denote the number of visual and query tokens, and $D$ is the dimension of hidden states. In this work, we adopt action tokens as query tokens. Via standard multi-head projection and reshaping, we derive query matrices $\mathbf{Q}^{\ell_s} \in \mathbb{R}^{N_h \times L_{\mathrm{query}} \times d}$ from query tokens and key matrices $\mathbf{K}^{\ell_s} \in \mathbb{R}^{N_h \times L_{\mathrm{vis}} \times d}$ from visual tokens, with $N_h$ being the number of attention heads and $d$ the dimension of each head.
The scaled dot-product attention $\mathbf{A}^{\ell_s}$ from query tokens to visual tokens is calculated as
\begin{equation}
\mathbf{A}^{\ell_s}=Softmax(\frac{\mathbf{Q}^{\ell_s}\big(\mathbf{K}^{\ell_s}\big)^{\top}}{\sqrt{d}})
\in\mathbb{R}^{N_h\times L_{\mathrm{query}}\times L_{\mathrm{vis}}}.
\end{equation}
Each entry $A^{\ell_s}_{h,t,v}$ measures how strongly the $t$-th query token attends to the $v$-th visual token under head $h$.
We aggregate attention across heads and query tokens to obtain a per-visual-token saliency score $\mathcal{S}^{\ell_s}$:
\begin{equation}
\mathcal{S}^{\ell_s}=\frac{1}{N_h\,L_{\mathrm{query}}}\sum_{h=1}^{N_h}\sum_{t=1}^{L_{\mathrm{query}}}\mathbf{A}^{\ell_s}_{h,t,v}\in\mathbb{R}^{L_{\mathrm{vis}}}.
\label{eq:saliency}
\end{equation}
Based on the score, we only keep the top-K highest-saliency visual tokens and prune the rest to focus computation on evidence most relevant to current task. It should be noted that the pruning operation is irreversible, meaning that only this subset of visual tokens is available for the subsequent inference phase. 
\input{figures/difference}
\subsection{Visual Attention Dynamics in VLA Models}
\label{sec:semantic_consistency}
The generation of action tokens at a given control timestep typically consists of multiple distinct stages. For instance, autoregressive models involve prefill and decoding stages, single-forward models perform inference across different depth of Transformer blocks, and two-stage models include separate semantic understanding and action generation phases. To achieve favorable acceleration efficiency, pruning is typically performed in the early stage of VLA inference. This raises a critical question: \textit{Will early-stage pruning prematurely discard visual tokens that the model will focus on in the late stages?} To investigate this issue, we thoroughly analyze how the attention patterns evolve across \textbf{intra-timestep stages} and \textbf{control timesteps}. Our key insights can be summarized as follows.

\noindent\textbf{Varied concentration across intra-timestep stages.}
As illustrated in Fig.~\ref{fig:difference}, we visualize the attention pattern differences between early and late stages based on the single-forward model OpenVLA-OFT~\cite{kim2025openvlaoft}. Early attention patterns are scattered and broad, while later attention patterns become more locally concentrated and focused, reflecting a coarse-to-fine attention regime. We further quantify how this discrepancy affects pruning. At pruning rates above 70\%, the probability that early-saliency-only pruning removes core tokens (the top 10\% most important tokens) of other layers rises sharply, causing substantial performance degradation in conventional pruning methods.

\input{figures/sac}
\noindent\textbf{Semantic consistency across control timesteps.}
Although attention patterns differ across intra-timestep stages, we observe that attention patterns at the same stage across different control timesteps, especially the more concentrated late-stage attention, exhibit striking semantic consistency. 
As depicted in Fig.~\ref{fig:sac}, the gripper moves from the middle to the right to grasp the soup and then to the left to place it; despite this large displacement, attention remains highly concentrated around the gripper region in every frame. This suggests that VLA models consistently attend to patches sharing the same task-critical \emph{semantics} throughout the episode, which in turn enables the future attention prediction strategy based on historical frames. It is worth noting that prior work~\cite{wang2025specprunevla,liu2026vlaprunertemporalawareduallevelvisual} has reported substantial overlap in attention patterns between adjacent timesteps, while our observation extends this finding to longer horizons and reveals a more general form of semantic consistency.

\noindent\textbf{Sudden changes at subtask transitions.}
Although semantic attention consistency holds for most steps, we sometimes observe abrupt shifts in attention at subtask transitions. For example, in Fig.~\ref{fig:sac}, frame~13 attends more to the basket compared with frame~12. This change corresponds to the transition from the ``pick up'' subtask to the ``place'' subtask, where the task-critical semantic entity changes from the object to the target box.

These observations jointly motivate our overall pruning framework. The varied attention patterns across intra-timestep stages demonstrate that effective pruning must consider the saliency of different stages to avoid the premature discarding of critical tokens of the late stage. However, since pruning is executed in the early stage, the late-stage saliency is unavailable at the pruning moment. To resolve this, we leverage the observed semantic consistency across control timesteps to predict the current step's late-stage saliency using information from prior steps, as introduced in Sec.~\ref{sec:future_forecast}. Furthermore, to mitigate unreliable predictions caused by the observed sudden attention shifts during subtask transitions as well as accumulated drift, we propose an adaptive reference timestep selection strategy, as introduced in Sec.~\ref{sec:subtask_division}. The pipeline of our full approach is illustrated in Fig.~\ref{fig:pipeline}.

\input{figures/pipeline}

\subsection{Future-Aware Token Pruning}

\label{sec:future_forecast}
Inspired by the semantic attention consistency introduced in Sec.~\ref{sec:semantic_consistency}, we propose a forward-looking strategy that \emph{forecasts} token importance in the late stages. 

Consider a VLA task with a total of T control timesteps. For each control timestep, the model generates the corresponding action tokens via forward propagation through $L$ Transformer layers. To extract representative timesteps for saliency prediction, we first select a subset of timesteps $\mathcal{T}_{\mathrm{key}}$ as \textbf{key timesteps}. For each key timestep $\tau \in \mathcal{T}_{\mathrm{key}}$, we do not prune at any layer and cache the full-token saliency vector
$\{\mathcal{S}^{\ell}_{\tau}, \ell=0,1,\dots,L-1\}$ for every layer $l$.
% $\{\mathcal{S}^{\ell}_{\tau}\}_{\ell=0}^{L-1}$. 

For each non-key timestep $t \notin \mathcal{T}_{\mathrm{key}}$, we perform pruning at the shallow layer $\ell_s$. We first calculate $\mathcal{S}^{\ell_s}_{t}$ as early-stage saliency. To predict the late stage saliency that we cannot directly compute at $\ell_s$, we take the most recent key timestep $\tau_{ref}$ as reference timestep, which is defined as:
\begin{equation}
\tau_{ref}=\max\{\tau\in\mathcal{T}_{\mathrm{key}}:\tau < t\}.
\label{eq:tau}
\end{equation}  
Afterwards, we identify the semantically closest visual token at the reference key timestep $\tau_{ref}$ for each visual token at the current timestep $t$, by minimizing the cosine distance between the corresponding tokens. Based on this visual token correspondence, we transfer the precomputed saliency scores $\mathcal{S}^{\ell}_{\tau_{ref}}$ of every layer $l$ from $\tau_{ref}$ to $t$, denoted as $\{\hat{\mathcal{S}}^{\ell}_{t}, \ell=0,1,\dots,L-1\}$.

To better approximate the importance of tokens in the late stages, we predict saliency vectors for a set of deeper layers $\mathcal{L}_d$ and aggregate them by averaging:
\begin{equation}
\hat{\mathcal{S}}^{\mathcal{L}_d}_{t}=\frac{1}{|\mathcal{L}_d|}\sum_{\ell\in\mathcal{L}_d}\hat{\mathcal{S}}^{\ell}_{t}\in\mathbb{R}^{L_{\mathrm{vis}}}.
\label{eq:forecast-avg}
\end{equation}

We then fuse the shallow-layer saliency and the deeper-layer predicted saliency to generate the final token importance scores:
\begin{equation}
\mathbf{s}_{t}= (1-\lambda)\,\mathcal{S}^{\ell_s}_{t} + \lambda\,\hat{\mathcal{S}}^{\mathcal{L}_d}_{t},
\qquad \lambda\in[0,1].
\label{eq:combine}
\end{equation}
Based on $\mathbf{s}_t$, we select the top-K visual tokens with the highest importance scores for retention. By jointly considering token importance in both current and future inference stages during pruning, our approach effectively mitigates the risk of prematurely discarding tokens that become critical in later stages, thereby achieving a better balance between performance and efficiency.

\subsection{Adaptive Key Timestep Selection}
\label{sec:subtask_division}
The key timestep set $\mathcal{T}_{\mathrm{key}}$ directly determines the quality of our future-saliency forecast and therefore plays a crucial role in the overall pruning performance.
A straightforward baseline is to sample key timesteps at a fixed temporal interval.
However, as discussed in Sec.~\ref{sec:semantic_consistency}, attention may change abruptly at subtask transitions; in such cases, an interval-based strategy can miss the transition point and propagate an outdated saliency prior across multiple consecutive steps, leading to suboptimal pruning performance.
We argue that the validity of the saliency prior is more correlated with whether two control timesteps belong to the \emph{same subtask} than with their temporal distance.
Motivated by this observation, we adaptively refresh key timesteps by detecting attention shifts via the cosine similarity of saliency maps.

We use the shallow-layer saliency vector $\mathcal{S}^{\ell_s}_{t}\in\mathbb{R}^{L_{\mathrm{vis}}}$ as a lightweight proxy of the current attention pattern and compare it with the forecasted saliency vector in the same layer. We compute the cosine similarity:
\begin{equation}
\kappa_t
=\frac{\left\langle \mathcal{S}^{\ell_s}_{t},\,\hat{\mathcal{S}}^{\ell_s}_{t}\right\rangle}
{\left\|\mathcal{S}^{\ell_s}_{t}\right\|_2\,\left\|\hat{\mathcal{S}}^{\ell_s}_{t}\right\|_2}
\in[-1,1].
\label{eq:saliency-cos}
\end{equation}
When $\kappa_t<\gamma$ (where $\gamma$ is a threshold), we detect an attention shift and refresh the key timestep:
\begin{equation}
t\in\mathcal{T}_{\mathrm{key}}\iff \kappa_t<\gamma.
\label{eq:key-select}
\end{equation}
At such a key timestep, we trigger full-token inference and update the stored saliency prior for subsequent forecasting; otherwise we reuse the latest key timestep $\tau_{ref}$ for saliency transfer and forecasting, followed by the token pruning.

Overall, at each timestep $t$, SAFE-Pruner: (i) computes shallow saliency $\mathcal{S}^{\ell_s}_{t}$; (ii) forecasts layer-wise saliency from the latest keyframe cache via Eq.~\eqref{eq:forecast-avg}; (iii) decides whether $t$ is a keyframe using Eqs.~\eqref{eq:saliency-cos}--\eqref{eq:key-select} in current shallow layer.
If $t\in\mathcal{T}_{\mathrm{key}}$, we \emph{disable pruning} and run full-token inference to directly output the action $\mathbf{a}_t$, while computing reliable saliency maps in different layers and updating the keyframe cache $\{\mathcal{S}^{\ell}_{\tau}\}$ for subsequent forecasting.
If $t\notin\mathcal{T}_{\mathrm{key}}$, we skip the expensive full-token computation and instead rely on the forecasted deep-layer saliency $\hat{\mathcal{S}}^{\mathcal{L}_d}_{t}$ (Eq.~\eqref{eq:forecast-avg}) together with the shallow metrics $\mathcal{S}^{\ell_s}_{t}$ to fuse scores and select tokens using Eq.~\eqref{eq:combine}, and run the original VLA policy on the pruned token set to output the action $\mathbf{a}_t$.
This yields a training-free, plug-and-play pruning pipeline that refreshes priors \emph{only when necessary} while preserving the accuracy benefits of future-aware saliency.

%% file: figures/difference.tex
\begin{figure}[tb]
  \centering
  \begin{subfigure}{0.50\linewidth}
    \centering
    \includegraphics[width=\linewidth]{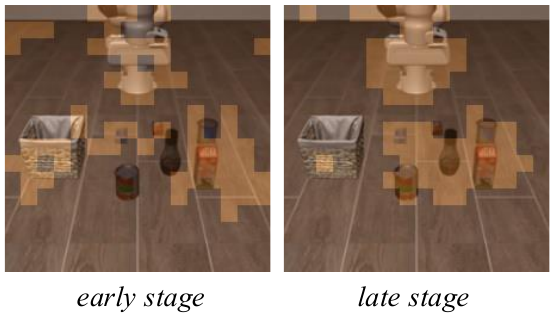}
    \caption{\small Saliency of different stages}
    \label{fig:difference-a}
  \end{subfigure}
  \hfill
  \begin{subfigure}{0.46\linewidth}
    \centering
    \includegraphics[width=\linewidth]{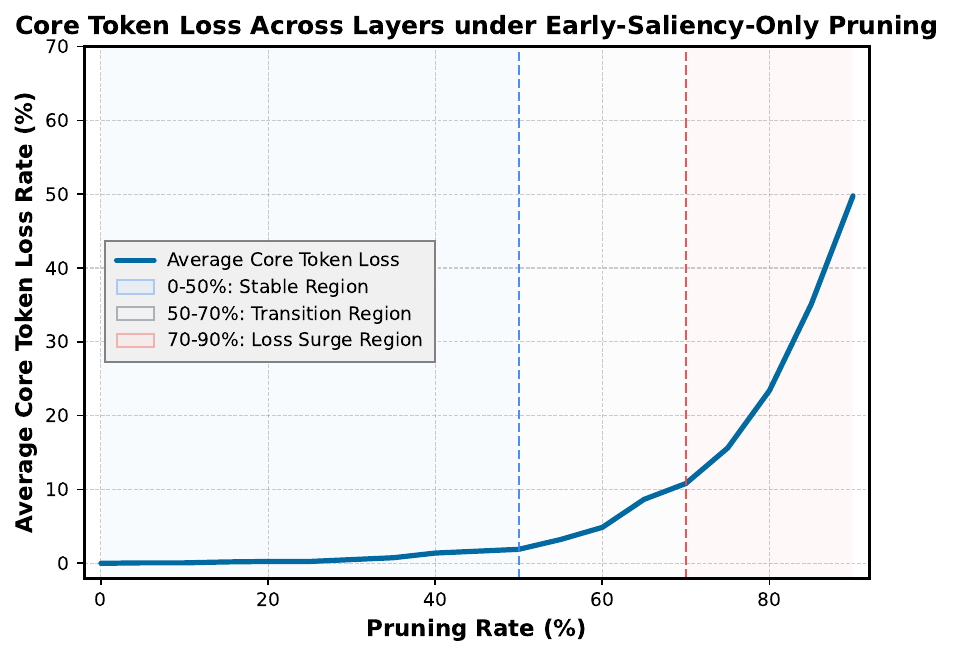}
    \caption{\small Core token loss rate}
    \label{fig:difference-b}
  \end{subfigure}
\caption{\textbf{Attention pattern differences and core token loss under early-saliency-only pruning in OpenVLA-OFT.} (a) Visualization of a coarse-to-fine attention focusing regime, where early layers show broad attention and late layers focus on critical regions. (b) Core token loss rate under early-saliency pruning, rising sharply at aggressive pruning levels.}
  \label{fig:difference}
\end{figure}

%% file: figures/sac.tex
\begin{figure}[tb]
  \centering
  \includegraphics[width=0.99\linewidth]{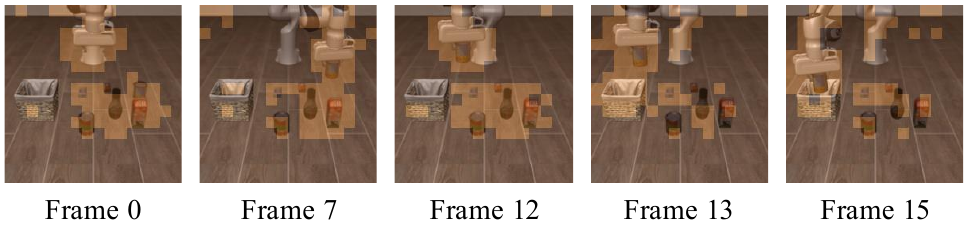}
\caption{\small \textbf{Visualization of semantic attention consistency and subtask transitions.} We visualize key attention tokens (orange) across five frames of a robotic pick-and-place episode. For most steps, attention remains anchored to task-relevant semantic entities rather than spatial positions, demonstrating semantic consistency. An abrupt shift in attention occurs at Frame 13, which coincides with the subtask transition. This temporal evolution highlights both the stability of semantic attention within a subtask and the abrupt shift at the boundary between two distinct subtasks.}

  \label{fig:sac}
\end{figure}

%% file: figures/pipeline.tex
\begin{figure}[tb]
  \centering
  \includegraphics[width=0.99\linewidth]{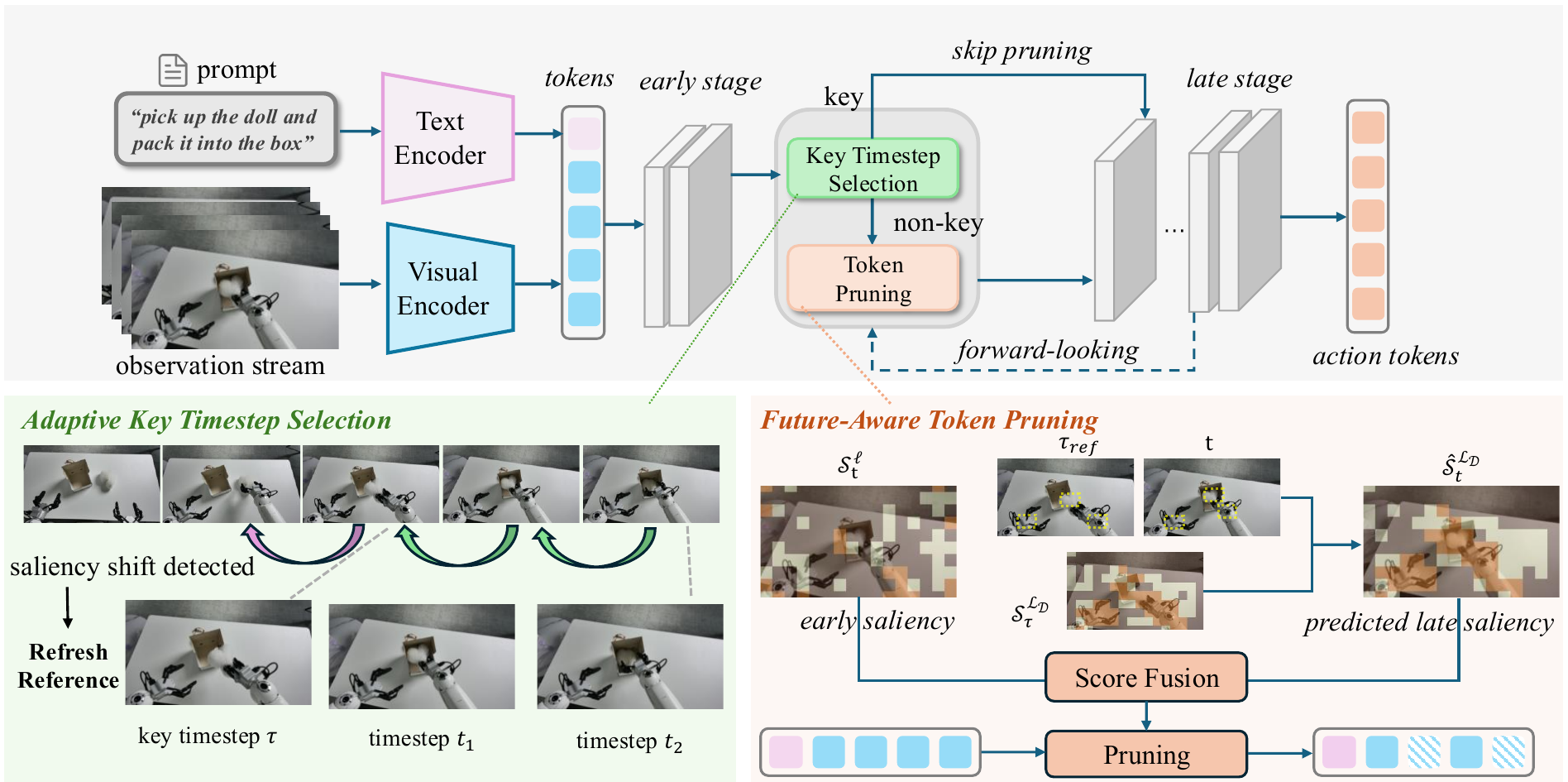}
  \caption{\small \textbf{The overall framework of SAFE-Pruner.} Given a text prompt and a sequence of visual frames, multimodal tokens are produced by text and visual encoders. A saliency-based keyframe detection module identifies frames with significant saliency shifts. For keyframes, pruning is skipped; for non-keyframes, our future-aware token pruning module predicts late-stage token saliency from historical frames, fuses it with early-stage saliency scores, and removes non-critical tokens. This forward-looking strategy incorporates pruning impact on future stages into decision, achieving a superior accuracy–efficiency balance in VLA inference.}
  \label{fig:pipeline}
\end{figure}

%% file: sections/4_experiment.tex
\section{Experiments}

\subsection{Experiment Setups}

\noindent\textbf{Foundation Models.} We implement our method on four VLA backbones with different architectures and decoding paradigms to test generalization. Specifically, we use: (1) OpenVLA~\cite{kim2024openvla}, a token-based VLA that predicts discretized actions conditioned on a fused vision encoder and an LLM backbone; (2) OpenVLA-OFT~\cite{kim2025openvlaoft}, which replaces step-by-step decoding with parallel decoding plus action chunking and continuous-action regression; (3) CogACT~\cite{li2024cogact}, a componentized VLA that separates vision/language reasoning from a specialized diffusion-based action module; and (4) $\pi_{0.5}$~\cite{intelligence2025pi05visionlanguageactionmodelopenworld}, which couples a VLM backbone with a flow-matching action expert. These models cover the mainstream architectural families in modern VLAs, and consistent gains across them verify our method’s cross-architecture generalization.
% (autoregressive token prediction, parallel decoding with chunked continuous actions, diffusion action modules, and flow-matching action experts)

\noindent\textbf{Comparison Methods.} We compare our approach against the unaccelerated vanilla model as primary baseline, together with a set of representative state-of-the-art open-source token pruning methods. These include VLM acceleration methods like FastV~\cite{chen2024fastv}, SparseVLM~\cite{zhang2024sparsevlm}, and DivPrune~\cite{alvar2025divprune}, as well as recent efficiency methods developed specifically for VLA models: VLA-Cache~\cite{xu2025vlacache} and VLA-Pruner~\cite{liu2026vlaprunertemporalawareduallevelvisual}. For all baseline methods, we conduct thorough hyperparameter tuning under the same experimental setup to obtain their best possible performance, ensuring a fair and convincing comparison.

% \noindent\textbf{Evaluation Metrics.} We use three evaluation metrics to jointly assess task performance and inference efficiency. Specifically, Success Rate is adopted to evaluate task completion quality. FLOPs is used to measure the theoretical computational complexity, and Latency is reported to characterize the actual inference speed of the VLM backbone. These metrics are widely adopted in current token pruning works, allowing us to comprehensively and fairly evaluate the performance-efficiency trade-off of different methods.

\noindent\textbf{Evaluation Metrics.} We use three evaluation metrics, success rate, FLOPs and latency, to jointly assess task performance and inference efficiency. These metrics are widely adopted in current token pruning works~\cite{xu2025vlacache,liu2026vlaprunertemporalawareduallevelvisual}, allowing us to comprehensively and fairly evaluate the performance-efficiency trade-off of different methods. For latency evaluation, following VLA-Cache~\cite{xu2025vlacache}, we report the per-timestep inference latency of the VLM backbone. Unless otherwise specified, all latency measurements are conducted on an NVIDIA RTX 4090 GPU. The reported latency is averaged over all evaluated control timesteps, including both key and non-key timesteps, and incorporates all additional computational overhead introduced by our method.

\subsection{Evaluation Benchmarks.}

\noindent\textbf{LIBERO~\cite{liu2023libero}.} We adopt the LIBERO benchmark as our primary evaluation suite, which assesses the generalization capability of VLA models across four manipulation dimensions: Spatial reasoning, Object-centric interaction, Goal-directed execution, and Long-horizon planning. Structured into four distinct suites, LIBERO features 10 unique manipulation tasks per suite, with 50 evaluation episodes allocated to each task—resulting in 2000 episodes. We conduct experiments on OpenVLA, OpenVLA-OFT and $\pi_{0.5}$ in this benchmark. To maintain consistent and strict experimental comparability with prior VLA work, we adhere to the standardized evaluation protocol and publicly released official weights for all baseline models.

\noindent\textbf{SIMPLER~\cite{li24simpler}.} We evaluate our method on the SIMPLER simulator to assess its robustness in bridging the simulation-to-reality gap. This simulator provides two complementary settings for a Google robot arm: Visual Matching and Variant Aggregation, each encompassing four distinct manipulation tasks. We select CogACT~\cite{li2024cogact} as our base model for this benchmark, as it provides official pre-trained weights tailored for SIMPLER, enabling us to reliably quantify the performance-efficiency trade-offs of different methods.

\input{figures/real_robot}

\noindent\textbf{Real Robot Experiment.} We conduct real-world evaluations using the Astribot S1~\cite{gao2025towards} dual-arm platform. With its chassis and torso immobilized, control is strictly limited to the 14-DoF arms and grippers. Visual observations are provided by two wrist-mounted RGB cameras alongside a head-mounted camera that actively tracks the workspace center. The detailed configuration is visualized in Fig.~\ref{fig:robot_setup}. Using 200 to 1,000 expert demonstrations collected via VR teleoperation, we fine-tune task-specific \(\pi_{0.5}\) models for three distinct scenarios designed to test a spectrum of robotic capabilities. These include \emph{Pick and Place} to assess basic manipulation across 10 different objects, \emph{Throw Basketball} to test small-object dexterity, and \emph{Pack Doll}---a long-horizon, multi-stage task requiring precise geometric coordination to pick up a doll, place it inside a box, and secure the lid. The task setups are visualized in Fig.~\ref{fig:real_robot}. Each trained model is evaluated across 200 trials to comprehensively measure its performance.

\input{tables/main-quant}
\input{tables/hif8}

\input{tables/simpler-quant}
\subsection{Main Results}
\noindent\textbf{Results on LIBERO.} The results of the LIBERO experiment are listed in Tab.~\ref{tab:libero-comparison}, which demonstrates the broad applicability and significant superiority of our method. Due to differences in model architecture and size, visual token pruning methods exhibit varying acceleration effects across different models. Nevertheless, our method consistently achieves an excellent performance-efficiency balance across diverse model architectures when compared to the unaccelerated vanilla model—incurring only a negligible performance drop of no more than 1.5\% in average success rate while reducing computational cost and latency by up to 60\% and 47\%, respectively. Furthermore, by integrating future attention cues into pruning decisions, our method maintains consistently superior performance over existing methods even under higher acceleration settings. This is evident in the OpenVLA-OFT~\cite{kim2025openvlaoft} experiment: while achieving substantially better acceleration in terms of FLOPs from 2.126 T to 1.722 T and backbone latency from 50.24 ms to 37.22 ms, our method still achieves the highest success rate, which is 1.9\% higher than the best performance of existing methods. These results confirm the superiority of our method in both efficiency and performance across different architectures of mainstream VLA models. We further evaluate the compatibility of SAFE-Pruner with HiF8~\cite{luo2024ascend} quantization,
as reported in Tab.~\ref{tab:hif8}. The combined framework reduces FLOPs by 56.4\% and GPU memory by 38.1\% relative to the vanilla OpenVLA-OFT model, while maintaining a comparable success rate. These results demonstrate that SAFE-Pruner is compatible with quantization methods, enabling simultaneous improvements in computational and memory efficiency.

\noindent\textbf{Results on SIMPLER.} We evaluate the sim-to-real robustness and efficiency of our method on the SIMPLER benchmark (Tab.~\ref{tab:simpler-comparison}), focusing on the CogACT backbone across two domain-shift settings: Visual Matching (VM) and Variant Aggregation (VA). Overall, our approach achieves state-of-the-art inference acceleration while preserving task performance comparable to the unpruned CogACT baseline, demonstrating its effectiveness in realistic robotic scenarios. In the VM setting, our method attains an average success rate of 74.5\%, nearly identical to the 74.8\% success rate achieved by the vanilla CogACT, while reducing FLOPs to just 37.4\% of the baseline and delivering a $1.73\times$ speedup. This outperforms all competing methods: FastV delivers only a $1.21\times$ speedup at 42.0\% FLOPs, while VLA-Cache offers a modest $1.38\times$ speedup at 80.1\% FLOPs. In the VA setting, our method maintains an average success rate of 61.9\%, which is on par with the baseline 61.3\%, and achieves a $1.67\times$ speedup with 36.2\% FLOPs retention, outperforming the $1.37\times$ speedup of VLA-Cache and the marginal acceleration of FastV. These results confirm that our token pruning strategy generalizes to sim-to-real scenarios, making it suitable for deployment on real robotic systems with strict latency constraints. In some settings, the accelerated model slightly outperforms the vanilla baseline, a phenomenon also observed in prior work~\cite{xu2025vlacache}. We assume that this is because our method filters out tokens irrelevant to the task, thus making the model more concentrated on the useful information and improving the success rate.

\noindent\textbf{Results on Real Robot.} We further validate our method in real-world deployment on the Astribot S1 dual-arm platform. We conduct a comprehensive evaluation of \textbf{FastV} and \textbf{Ours}, reporting task success rate together with computational cost and VLM backbone inference latency measured on an NVIDIA RTX 3090. The results are shown in Tab.~\ref{tab:real_robot}. Compared with the unpruned \(\pi_{0.5}\) baseline, both pruning methods substantially reduce inference cost and latency, decreasing FLOPs from 2.264 T to 1.857 T with FastV and to 1.543 T with our method, while reducing latency from 80.36 ms to 56.93 ms and 43.56 ms, respectively. This efficiency gain comes with a drop in average success rate from 81.3\% to 69.0\% for FastV and to 79.3\% for our method, showing that our approach preserves performance better. Overall, our method achieves a consistently better accuracy--efficiency trade-off than FastV. The gains are most pronounced on \textit{Throw Basketball} from 69\% to 81\%, which requires precise grasping of small objects. We also observe improvements from 82\% to 90\% on \textit{Pick and Place} and from 56\% to 67\% on the long-horizon \textit{Pack Doll} task. These results indicate that incorporating future-aware attention cues yields more reliable pruning decisions in real robotic control, preserving task-critical visual evidence while reducing inference cost.

\input{tables/real_robo-quant}

\input{tables/ablation-quant}
\subsection{Analysis}
To validate the effectiveness of each component in our method, we perform systematic ablation studies using OpenVLA-OFT~\cite{kim2025openvlaoft} on the LIBERO benchmarks~\cite{liu2023libero}, evaluating four experimental configurations: (1) the unaccelerated model (vanilla), (2) token pruning only using shallow attention without future state forecasting (w/o forecast), (3) future-layer prediction guided pruning with fixed-interval keyframe selection (w/o adaptivity) and (4) our complete pipeline (Ours). As presented in Tab.~\ref{tab:ablation-study}, the vanilla baseline achieves a high average success rate of 96.8\% but introduces heavy computational costs with 3.970 T FLOPs and 70.25 ms latency; although w/o forecast delivers obvious acceleration, it suffers from a clear performance drop to 94.5\% due to the premature discarding of task-relevant tokens. By incorporating future-layer information to assist token pruning, w/o adaptivity alleviates accuracy degradation and boosts the average success rate to 95.8\%, but it fails to account for attention shift at subtask boundaries, and full-token inference on fixed-interval keyframes limits further acceleration. In contrast, our full approach incorporates an adaptive key-timestep selection mechanism, which improves prediction robustness while reducing the number of key timesteps. As a result, it achieves the highest average success rate of 96.4\%, along with the lowest FLOPs of 1.722 T and latency of 37.22 ms among acceleration variants, demonstrating that future forecasting and adaptive reference refreshing jointly balance performance and inference efficiency.

%% file: figures/real_robot.tex
\begin{figure}[tb]
  \centering
  \includegraphics[width=\linewidth]{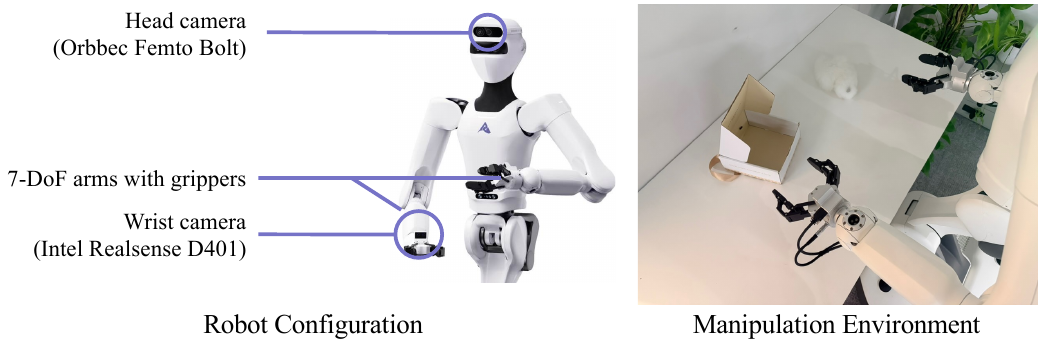}
  % \caption{\small \textbf{Visualization of the real-world experiment setup.}}
\caption{\small \textbf{Visualization of the real-world experiment setup.} Left: The Astribot S1 dual-arm platform, equipped with 7-DoF arms, grippers, and wrist-mounted and head-mounted RGB cameras for visual perception. Right: The manipulation environment used to evaluate task performance across diverse robotic scenarios.}
  \label{fig:robot_setup}
\end{figure}

\begin{figure}[tb]
  \centering
  \includegraphics[width=0.99\linewidth]{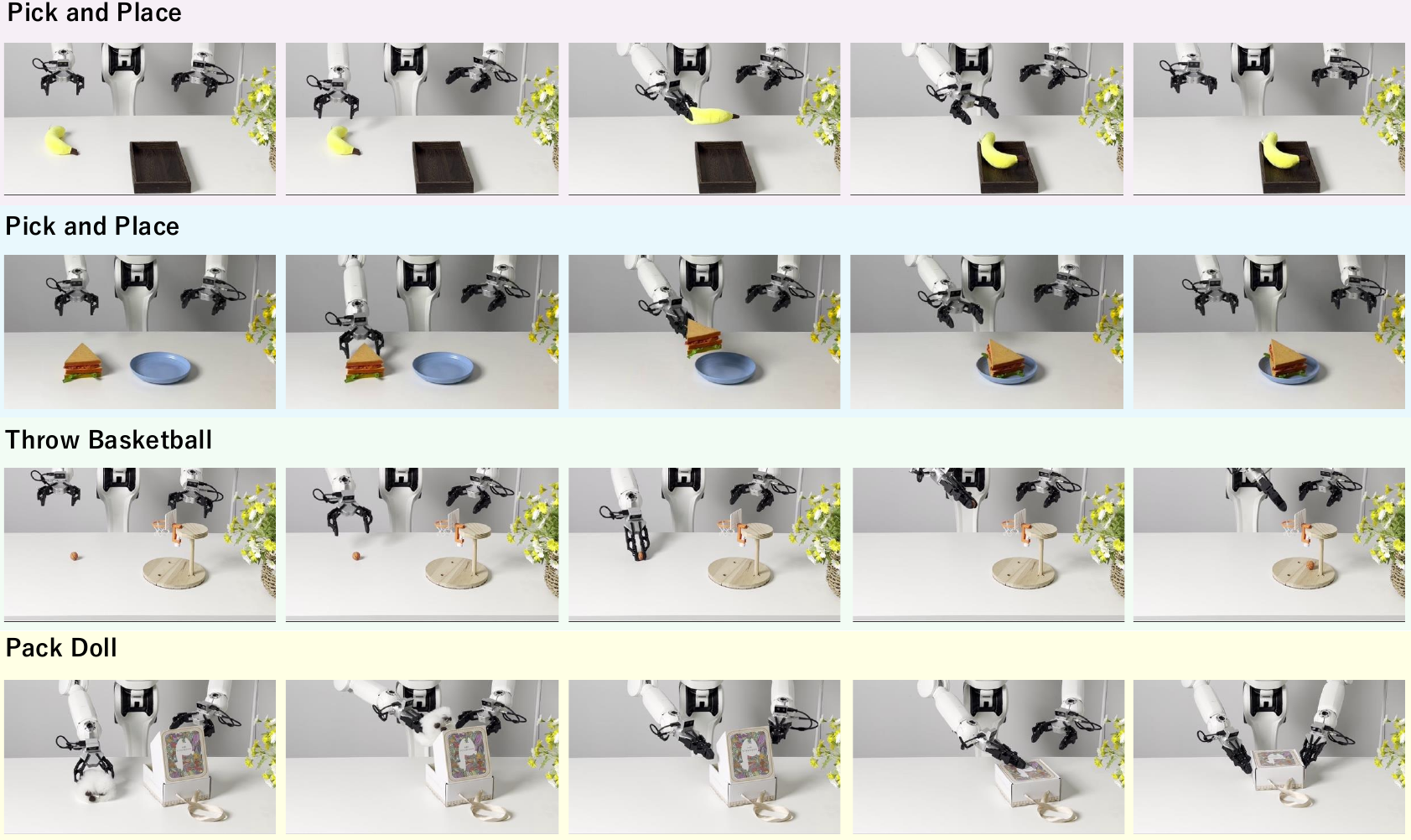}
\caption{\small \textbf{Real-world robotic task scenarios for evaluation.} Visualization of three distinct task scenarios on the Astribot S1 platform: Pick and Place task for basic manipulation, Throw Basketball for small-object dexterity, and Pack Doll for long-horizon multi-stage coordination, each designed to assess different robotic capabilities.}
  \label{fig:real_robot}
\end{figure}

%% file: tables/main-quant.tex
\begin{table}[tb]
\centering
% \caption{Comparison on LIBERO benchmarks.}
\caption{\small \textbf{Comparison on LIBERO benchmarks.} Best results among acceleration methods are highlighted in \textbf{bold}. Our method consistently achieves the best trade-off between efficiency and performance across models with different architectures, delivering the lowest FLOPs and latency while maintaining top success rates.}
\label{tab:libero-comparison}
\begin{tabular}{l l c c c c c c c} % 9 columns: Model, Method, 5 success cols, FLOPs, Latency
\toprule

% \multirow{2}{*}{Model} & \multirow{2}{*}{Method} & \multicolumn{5}{c}{Success Rate $\uparrow$} & \multirow{2}{*}{FLOPs (T)$\downarrow$} & \multirow{2}{*}{Latency (ms)$\downarrow$} \\
% \cmidrule(lr){3-7}

\multirow{2}{*}{Model} & \multirow{2}{*}{Method} 
& \multicolumn{5}{c}{Success Rate(\%) $\uparrow$} 
& \multirow{2}{*}{\makecell{FLOPs\\(T)$\downarrow$}} 
& \multirow{2}{*}{\makecell{Latency\\(ms)$\downarrow$}} \\
\cmidrule(lr){3-7}

 &  & Spatial & Object & Goal & Long & Average & & \\
\midrule

\multirow{7}{*}{OpenVLA} 
 & Vanilla     & 84.6 & 86.6 & 78.2 & 53.2 & 75.7 & 1.862 & 48.24 \\
 & FastV       & 81.8 & 70.6 & 71.8 & 44.6 & 67.2 & 0.811 & 34.63 \\
 & SparseVLM   & 80.6 & 72.4 & 68.2 & 46.6 & 67.0 & 0.876 & 34.32 \\
 & DivPrune   & 78.6 & 62.8 & 66.2 & 43.2 & 62.7 & 0.832 & 37.18 \\
 & VLA-Cache   & 78.2 & 71.2 & 70.6 & 45.8 & 66.5 & 0.941 & 35.21 \\
 & VLA-Pruner  & 82.0 & 84.4 & \textbf{77.8} & \textbf{52.6} & 74.2 & 0.793 & 36.47 \\
 & Ours        & \textbf{82.2} & \textbf{84.8} & 77.6 & 52.0 & \textbf{74.2} & \textbf{0.742} & \textbf{32.18} \\
\midrule

\multirow{7}{*}{OpenVLA-OFT} 
 & Vanilla     & 98.4 & 98.2 & 96.4 & 94.2 & 96.8 & 3.970 & 70.25 \\
 & FastV       & 97.8 & 92.8 & 95.6 & 91.8 & 94.5 & 2.141 & 50.24 \\
 & SparseVLM  & \textbf{98.6} & 91.8 & 94.8 & 92.0 & 94.3 & 2.289 & 52.15 \\
 & DivPrune    & 97.2 & 90.0 & 93.4 & 89.6 & 92.6 & 2.126 & 53.87 \\
 & VLA-Cache   & 97.0 & 92.2 & 94.6 & 91.0 & 93.7 & 2.730 & 58.14 \\
 & VLA-Pruner  & 97.4 & 93.0 & 94.0 & 92.0 & 94.1 & 2.234 & 54.53 \\
 & Ours        & 98.0 & \textbf{98.0} & \textbf{96.2} & \textbf{93.4} & \textbf{96.4} & \textbf{1.722} & \textbf{37.22} \\
\midrule

\multirow{7}{*}{$\pi_{0.5}$}
 & Vanilla     & 98.2 & 98.0 & 98.0 & 91.8 & 96.5 & 2.115 & 35.28 \\
 & FastV      & 96.4 & 96.8 & 95.6 & 89.0 & 94.5 & 1.612 & 28.64 \\
 & SparseVLM   & 94.6 & 97.2 & 94.8 & 89.2 & 94.0 & 1.713 & 29.03 \\
 & DivPrune    & 93.8 & 96.6 & 92.4 & 88.0 & 92.7 & 1.576 & 30.55 \\
 & VLA-Cache   & 94.8 & 96.0 & \textbf{96.4} & 88.2 & 93.9 & 1.632 & 28.14 \\
 & VLA-Pruner  & 95.8 & 97.2 & 95.6 & 88.0 & 94.2 & 1.583 & 32.64 \\
 & Ours        & \textbf{97.0} & \textbf{98.8} & 96.0 & \textbf{90.2} & \textbf{95.5} & \textbf{1.482} & \textbf{24.33} \\

\bottomrule
\end{tabular}
\end{table}

%% file: tables/hif8.tex
% \begin{table}[tb]
% \centering
% \caption{\small
% \textbf{Comparison on LIBERO benchmarks.}
% Best results among acceleration methods are highlighted in \textbf{bold}.
% Our method consistently achieves the best trade-off between efficiency and
% performance across models with different architectures, delivering the lowest
% FLOPs and latency while maintaining top success rates.
% }
% \label{tab:hif8}

% % \setlength{\tabcolsep}{5pt}
% % \renewcommand{\arraystretch}{1.15}

% \begin{tabular}{@{}l c c c c c c c@{}}
% \toprule
% \multirow{2}{*}{Method}
% & \multicolumn{5}{c}{Success Rate (\%) $\uparrow$}
% & \multirow{2}{*}{\makecell{FLOPs\\(T) $\downarrow$}}
% & \multirow{2}{*}{\makecell{GPU Memory\\(GB) $\downarrow$}} \\
% \cmidrule(lr){2-6}
% & Spatial & Object & Goal & Long & Average & & \\
% \midrule

% Vanilla
% & 98.4 & 98.2 & 96.4 & 94.2 & 96.8
% & 3.970 & 17.83 \\

% Ours
% & 98.0 & 98.0 &96.2 &93.4 &96.4 
% &1.722 & 18.31 \\

% Ours + HiF8
% & 98.8 & 97.8 & 90.0 & 89.2 & 94.0
% & 1.718 & 10.44 \\

% Ours + HiF8 (VLM only)
% & 98.8 & 98.6 & 95.8 & 93.4 & 96.6
% & 1.729 & 11.03 \\
% \bottomrule
% \end{tabular}
% \end{table}

\begin{table}[tb]
\centering
\caption{\small
\textbf{Compatibility with quantization.} Combining SAFE-Pruner with HiF8 jointly reduces computational cost and GPU memory while maintaining performance comparable to the vanilla OpenVLA-OFT model.
}
\label{tab:hif8}

\begin{tabular}{@{}l c c c c c c c@{}}
\toprule
\multirow{2}{*}{Method}
& \multicolumn{5}{c}{Success Rate (\%) $\uparrow$}
& \multirow{2}{*}{\makecell{FLOPs\\(T) $\downarrow$}}
& \multirow{2}{*}{\makecell{GPU Memory\\(GB) $\downarrow$}} \\
\cmidrule(lr){2-6}
& Spatial & Object & Goal & Long & Average & & \\
\midrule

Vanilla
& 98.4 & 98.2 & 96.4 & 94.2 & 96.8
& 3.970 & 17.83 \\

Ours
& 98.0 & 98.0 &96.2 &93.4 &96.4 
&1.722 & 18.31 \\

Ours + HiF8
& 98.8 & 98.6 & 95.8 & 93.4 & 96.6
& 1.729 & 11.03 \\
\bottomrule
\end{tabular}
\end{table}

%% file: tables/simpler-quant.tex
\begin{table}[tb]
\centering
% \caption{Comparison on SIMPLER benchmarks.}
\caption{\small \textbf{Comparison on SIMPLER benchmarks.} Best results among acceleration methods are highlighted in \textbf{bold}. The results demonstrate that our approach yields the highest speedup and lowest computational cost, while maintaining a success rate competitive with or better than the baseline and other acceleration methods.}

\label{tab:simpler-comparison}
\begin{tabular}{l l c c c c c c c} % 9 columns: Model, Method, 5 success cols, FLOPs, Latency
\toprule

\multirow{2}{*}{SIMPLER} & \multirow{2}{*}{Method} 
& \multicolumn{5}{c}{Success Rate(\%) $\uparrow$} 
& \multirow{2}{*}{\makecell{FLOPs\\(\%)$\downarrow$}} 
& \multirow{2}{*}{\makecell{Speed\\up$\uparrow$}} \\
\cmidrule(lr){3-7}

 &  & Pick & Move & Drawer & Apple & Average & & \\
\midrule

\multirow{5}{*}{\makecell{Visual\\Matching}} 
 & CogACT       & 91.3 & 85.0 & 71.8 & 50.9 & 74.8 & 100.0 & $1.00\times$ \\
 & Random       & 9.7  & 20.4 & 53.5 & 0.0  & 20.9 & 58.5  & $1.20\times$ \\
 & FastV        & \textbf{92.6} & 81.4 & 69.8 & \textbf{52.4} & 74.1 & 42.0  & $1.21\times$ \\
 & VLA-Cache    & 92.0 & 83.3 & 70.5 & 51.6 & 74.4 & 80.1 & $1.38\times$ \\
 % & EfficientVLA & 93.3 & 81.3 & 68.2 & 53.8 & 74.2 & 28.9 & $1.93\times$ \\
 & Ours         & 91.3 & \textbf{83.8} & \textbf{70.8} & 51.9 & \textbf{74.5} & \textbf{37.4} & $\mathbf{1.73\times}$ \\
\midrule

\multirow{5}{*}{\makecell{Variant\\Aggregation}} 
 & CogACT        & 89.6 & 80.8 & 28.3 & 46.6 & 61.3 & 100.0 & $1.00\times$ \\
 & Random        & 4.0  & 16.1 & 15.6 & 0.0  & 8.9  & 58.5  & $1.20\times$ \\
 & FastV         & 91.4 & 78.6 & 27.6 & \textbf{50.6} & 62.1 & 42.0  & $1.19\times$ \\
 & VLA-Cache     & 91.7 & \textbf{79.3} & \textbf{32.5} & 45.8 & \textbf{62.3} & 82.6  & $1.37\times$ \\
 % & EfficientVLA  & 93.2 & 75.8 & 26.9 & 49.2 & 61.2 & 28.9  & $1.91\times$ \\
 & Ours          & \textbf{92.1} & 79.2 & 28.0 & 48.1 & 61.9 & \textbf{36.2} & $\mathbf{1.67\times}$ \\

\bottomrule
\end{tabular}
\end{table}

%% file: tables/real_robo-quant.tex
\begin{table}[t]
    \centering
    \caption{\small \textbf{Comparison on real robot manipulation tasks.}
    `Latency' refers to the inference latency of the VLM backbone,
    tested on an NVIDIA RTX 3090 GPU.}
    \label{tab:real_robot}

    \small
    \setlength{\tabcolsep}{3.5pt}
    \renewcommand{\arraystretch}{1.05}

    \begin{tabular}{@{}lccccccc@{}}
    \toprule
    \multirow{3}{*}{Method}
    & \multicolumn{4}{c}{Success Rate $\uparrow$}
    & FLOPs
    & Latency \\
    \cmidrule(lr){2-5}
    & \makecell{\textit{Pick}\\\textit{and Place}}
    & \makecell{\textit{Throw}\\\textit{Basketball}}
    & \makecell{\textit{Pack}\\\textit{Doll}}
    & \textit{Average}
    & (T)$\downarrow$
    & (ms)$\downarrow$ \\
    \midrule
    $\pi_{0.5}$
    & \textbf{94\%}
    & 77\%
    & \textbf{73\%}
    & \textbf{81.3\%}
    & 2.264
    & 80.36 \\

    \textbf{+ FastV}
    & 82\%
    & 69\%
    & 56\%
    & 69.0\%
    & 1.857
    & 56.93 \\

    \textbf{+ Ours}
    & 90\%
    & \textbf{81\%}
    & 67\%
    & 79.3\%
    & \textbf{1.543}
    & \textbf{43.56} \\
    \bottomrule
    \end{tabular}
\end{table}

%% file: tables/ablation-quant.tex
\begin{table}[tb]
\centering
% \caption{Ablation Studies.}
\caption{\small \textbf{Ablation Studies.} The ablation results verify the effectiveness of each proposed component, showing that both the forecasting mechanism and the keyframe selection strategy contribute to improved performance–efficiency balance, and their combination yields the best overall results.}
\label{tab:ablation-study}
\begin{tabular}{l c c c c c c c} % 8 columns: Method, 5 success cols, FLOPs, Latency
\toprule
\multirow{2}{*}{Method} 
& \multicolumn{5}{c}{Success Rate(\%) $\uparrow$} 
& \multirow{2}{*}{\makecell{FLOPs\\(T)$\downarrow$}} 
& \multirow{2}{*}{\makecell{Latency\\(ms)$\downarrow$}} \\
\cmidrule(lr){2-6}  
& Spatial & Object & Goal & Long & Average & & \\
\midrule
Vanilla       & 98.4 & 98.2 & 96.4 & 94.2 & 96.8 & 3.970 & 70.25 \\
w/o forecast  & 97.8 & 93.0 & 95.8 & 91.6 & 94.5 & 2.141 & 50.31 \\
w/o adaptivity  & \textbf{98.2} & 96.2 & 96.0 & 92.6 & 95.8 & 2.236 & 51.98 \\
Ours          & 98.0 & \textbf{98.0} & \textbf{96.2} & \textbf{93.4} & \textbf{96.4} & \textbf{1.722} & \textbf{37.22} \\
\bottomrule
\end{tabular}
\end{table}

%% file: sections/5_conclusion.tex
\section{Conclusion}
We present SAFE-Pruner, a training-free, plug-and-play token pruning framework that leverages the observed phenomenon, semantic attention consistency, to forecast late-stage token saliency from historical keyframes and thus avoid premature removal of task-critical visual evidence. By fusing shallow-stage cues with predicted deep-stage saliency and adaptively selecting key timesteps, our method successfully preserves downstream reasoning while substantially reducing computation and inference delay. Experiments across multiple VLA backbones, simulation benchmarks and real-robot trials demonstrate that SAFE-Pruner consistently improves the accuracy–efficiency trade-off, achieving up to $1.89\times$ speedup with negligible loss in task success. We believe SAFE-Pruner not only offers a practical path toward low-latency VLA inference for real-time robotic control, but also provides new insights into the attention dynamics of VLA models.